\documentclass[conference]{IEEEtran}

\usepackage{graphicx}

\usepackage{amsmath}
\usepackage{amssymb}
\usepackage{comment}
\usepackage{epsfig} 
\usepackage{subcaption}
\usepackage{caption}
\usepackage{placeins}
\usepackage{csquotes}

\usepackage{duckuments}
\usepackage{algorithm}
\usepackage{tabularx}
\usepackage{url}
\usepackage{accents}
\usepackage{pgfplots}
\pgfplotsset{compat=1.18}
\usepackage{lipsum}
\usepackage{xcolor}
\usepackage{enumitem}
\usepackage{multirow}

\usepackage{tikz}
\usepackage{graphicx}
\usetikzlibrary{positioning}

\begin{document}

\title{Learning Manipulation Tasks in Dynamic and Shared 3D Spaces}

\author{\IEEEauthorblockN{Hariharan Arunachalam}
\IEEEauthorblockA{
\textit{University of Lincoln}\\
United Kingdom \\
26760953@students.lincoln.ac.uk}
\and
\IEEEauthorblockN{Marc Hanheide}
\IEEEauthorblockA{
\textit{University of Lincoln}\\
United Kingdom \\
mhanheide@lincoln.ac.uk}
\and
\IEEEauthorblockN{Sariah Mghames}
\IEEEauthorblockA{
\textit{University of Lincoln}\\
United Kingdom\\
smghames@lincoln.ac.uk}
}

\maketitle

\begin{abstract}
Automating the segregation process is a need for every sector experiencing a high volume of materials handling, repetitive and exhaustive operations, in addition to risky exposures. Learning automated pick-and-place operations can be efficiently done by introducing collaborative autonomous systems (e.g. manipulators) in the workplace and among human operators. In this paper, we propose a deep reinforcement learning strategy to learn the place task of multi-categorical items from a shared workspace between dual-manipulators and to multi-goal destinations, assuming the pick has been already completed. The learning strategy leverages first a stochastic actor-critic framework to train an agent's policy network, and second, a dynamic 3D Gym environment where both static and dynamic obstacles (e.g. human factors and robot mate) constitute the state space of a Markov decision process. Learning is conducted in a Gazebo simulator and experiments show an increase in cumulative reward function for the agent further away from human factors. Future investigations will be conducted to enhance the task performance for both agents simultaneously.
\end{abstract}


\section{Introduction}

\begin{figure}[t]
    \centering
    \begin{subfigure}[b]{0.48\columnwidth}
    \centering
    \includegraphics[width=4cm , trim={0 1cm 0 1cm} , clip]{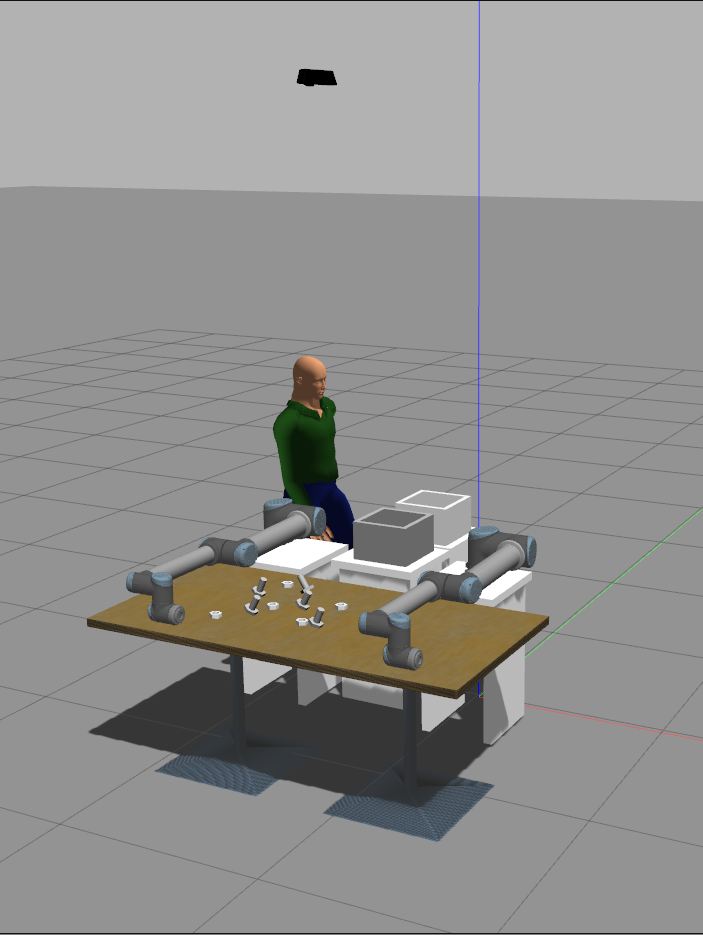}
    \caption{Side-view}
    \label{fig:side-view}
    \end{subfigure}
    \hfill
    \begin{subfigure}[b]{0.48\columnwidth}
    \centering
    \includegraphics[width=4cm , trim={0 0cm 0 0.7cm} , clip]{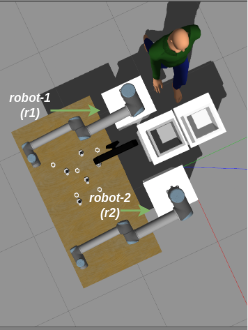}
    \caption{Top-view}
    \label{fig:top-view}
    \end{subfigure}
    \hfill
    
    \caption{Gazebo simulation environment to train a reinforcement learning agent for the autonomous collaborative segregation task. In the environment, two UR10, bolts and nuts items on a shared work table, two destination boxes, a mobile agent, and a Kinect sensor, are loaded.}
    \label{fig:scenario}
\end{figure}

The process of waste or item segregation though is essential for multiple applications, it can be in parallel time-demanding and exhaustive in view of the repetitive nature of the task. The latter is true when items being segregated are for example left-overs items from warehouses and manufacturing industries (e.g bolts and nuts), multi-categorical waste (e-waste, sanitary waste, nuclear waste, dry waste, wet waste, plastic waste, paper waste, etc). In this work, we propose to approach the autonomous segregation task from a collaborative perspective in order to optimize the processing speed when a large heap of multi-categorical items is collected and delivered to a segregation point. More specifically, given a single set of destination points and a shared heap of mix-categorical items that need to be sorted out into the set, as can be seen in Fig.~\ref{fig:scenario}, the problem now becomes to autonomously, efficiently and safely segregate single items into their corresponding destination, while avoiding static and dynamic obstacles in the surrounding, including for e.g. the mate autonomous system (robotic manipulator), operators nearby for e.g. for maintenance or to replace a filled destination point. Hence, the advantages of introducing collaborative autonomous systems to the segregation process can be summarised into the following:
\begin{itemize}
\item Speeding up the segregation process
\item Shorter usage of the hardware lifetime cycle. Consequently, less hardware failure can be expected.
\item When one robot fails and undergoes maintenance, the operation will not stop but will be pursued by the other agent in the same workspace.
\end{itemize}

Learning pick-and-place tasks is a common research problem that has been approached to date differently. 
Some works proposed optimization-based methods to plan safely the trajectory of a single manipulator in dynamic environments \cite{mikestilman, HILrafael2020, OptIProMP}. 
Other works proposed primitive-based methods to plan efficient online trajectories~\cite{FrankPSC22, mghames2020interactive}.
While other works, proposed more advanced techniques, relying on deep reinforcement learning to learn the pick or pick-and-place task~\cite{affordancemap2023, BEJJANI2021103730, gu2017deep}.

Though optimization and imitation-based methods are fast for real-world operations and some can adapt to dynamic environments, optimization-based methods rely on accurate system models.
The ability of a reinforcement learning (RL) framework to adapt the learned task to new and unknown dynamics in an end-to-end exploration fashion makes it an efficient approach to learning the segregation task in dynamic and shared workspace. While most of the aforementioned works deal with a single autonomous agent learning a task in 2D or 3D environments, the only works dealing with collaborative or cooperative multi-agents are \cite{gu2017deep, everett2018motion, prianto2021deep}. This is in addition to the coordinated robots introduced in \cite{icra23coordination} for pick-and-place tasks in cluttered and confined environments. In addition, learning other types of tasks (e.g. navigation) between collaborative agents has been proposed in \cite{everett2018motion}. Although the latter works show promising results, they do not deal with the specific problem of shared collaborative workspace, where each agent should be aware of the mate one. 
In this paper, we propose the MLDEnv framework (Manipulation Learning in Dynamic ENVironments) in which we approach the problem of learning pick-and-place segregation tasks in dynamic and shared workspace between multi-agents, in two stages: a first stage leverages a PointNet architecture to extract features from the environment, and a second stage leverages a PPO reinforcement learning technique to efficiently and safely generalize the task to different real-world environments. We release the work implementation as open-source\footnote {https://github.com/hariharan20/MLDEnv}.

\section{Approach} \label{sec:appr}
\graphicspath{{images/}}
 \subsection{Problem Definition}
In the process of making robotic manipulators environment-aware, several works have been done in the past decades. Implementation of conventional motion planners is not suitable for dynamic scenarios where real-time computational cost is high and convergence speed is of paramount importance. These planners are optimization-based problem solvers and are slower than data-driven techniques. A data-driven or machine learning planner deals with an approximated mathematical function mapping the observable states to the actions of the robot. However, there are past works carried out for manipulators task learning with dynamic obstacle avoidance as~\cite{prianto2021deep}. The latter work, for example, poses some limitations given the motion pattern of the dynamic obstacles involved and, the constrained (i.e. non-shared) workspace assigned to each robot. Though the latter works learn a manipulation task (e.g. pick-and-place) in 3D dynamic environments, they don't address the situations of both dynamic and shared 3D spaces between multi-agents. In this paper, we propose a two-stage data-driven framework for multi-agent manipulators task learning under 3D dynamic and shared environments. The framework uses 3D spatial data for perception to learn a representation of the environment.
\subsection{Scenario}
We consider a pick and place manipulation scenario where two base-fixed manipulators pick an object from a shared workspace (e.g. table) and place it in different boxes with shared access to each. This scenario applies to real-world problems where a heap of multi-categorical waste needs sorting for energy-efficient recycling, and a heap of left-over industrial items, for example, screws and nuts of different sizes, are picked up by an operator or autonomous system, at the end of a day, and thrown on a shared table for automated sorting and tidying. To test the proposed framework, the environment was created in simulation to train an agent policy. The scene involves also a human operator moving towards and away from the system, in predefined linear path segments at varying speeds, to replace the boxes. Human operators act therefore as obstacles to robots. Our case study consists of two different objects to be sorted in two different boxes as can be seen from Fig.~\ref{fig:scenario}. The scene also consists of a depth camera placed on top of the setup to capture the 3D point cloud of the scene. The whole scene was set up in the Gazebo simulator, and 3D objects were designed in Blender. 

\FloatBarrier
For reference purposes, the robotic manipulator near the human agent will be denoted as \textit{robot-1} while the other manipulator will be denoted as \textit{robot-2}. We assign the grey (near the table) box as a goal to \textit{robot-1}, while the white (farther from the table) box is the goal for \textit{robot-2}. The robots used in the simulation are UR10, a six degrees of freedom manipulator from Universal Robots. The robots are configured to receive velocity commands for movement.
\subsection{Perception}

\begin{figure}[t]
    \centering
    \begin{subfigure}[b]{0.48\columnwidth}
    \centering
    \includegraphics[width=4cm]{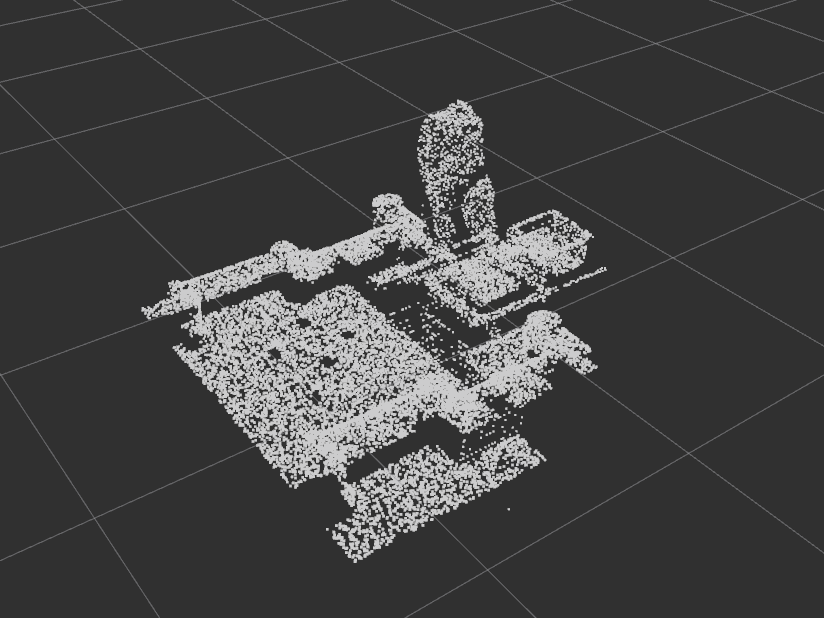}
    \caption{Input Point Cloud}
    \label{fig:input-pcd}
    \end{subfigure}
    \hfill
    \begin{subfigure}[b]{0.48\columnwidth}
    \centering
    \includegraphics[width=4cm]{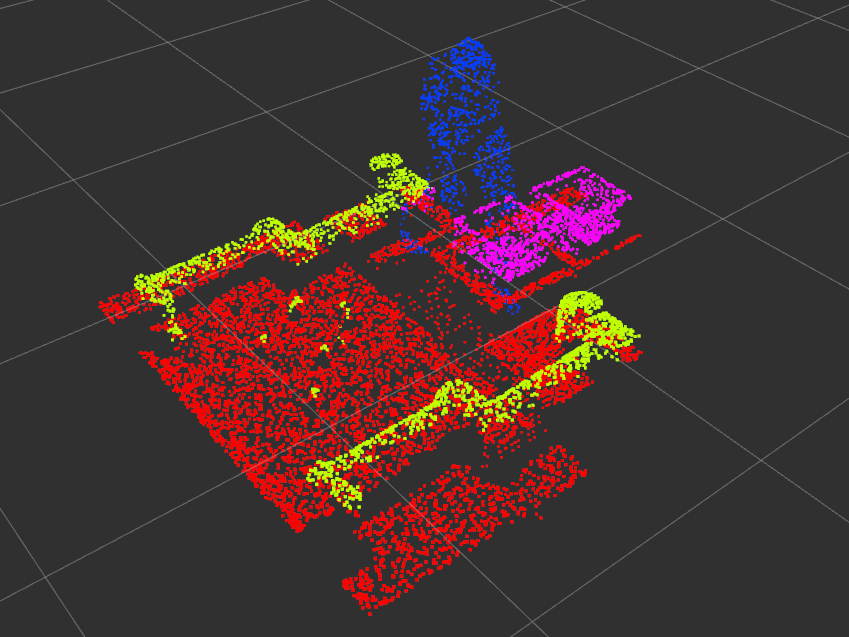}
    \caption{Output Point Cloud}
    \label{fig:output-pcd}
    \end{subfigure}
    \hfill
    
    \caption{Segmentation results of a pointcloud from the trained pointnet.}
    \label{fig:pointnet-inference}
\end{figure}

\textbf{Scene Segmentation.}
We train a 3D point cloud-based segmentation framework, the \textit{PointNet}~\cite{qi2017pointnet}, on a custom dataset to perform semantic segmentation of the scene. The point cloud dataset was collected from the simulation environment and was annotated for individual objects.   
The classes created for semantic segmentation were:
(a) table,
(b) human, 
(c) box,
(d) robot,
(e) cube.
The \textit{PointNet} produces a classification score for each point in the point cloud, as in the example given in Fig.~\ref{fig:pointnet-inference}. The points that are classified as ‘human’ are considered obstacles, these points are then fed into a DBSCAN (Density-Based Spatial Clustering of Applications with Noise), an unsupervised machine learning technique for clustering point clouds based on density. The resultant clusters are considered individual obstacles for each robot. We note that this paper doesn't present a performance comparison between 3D perception architectures on the robot task learning, hence a more modern 3D point-cloud based segmentation model could have been used instead to drive the task learning pipeline, as the work in~\cite{yang2023sam3d}.

Since some of the six joints of the selected manipulator are close to each other, these are reduced to four joints in the environment state space and are used in the measure of the distances to obstacles. Hence, from each of the resulting clusters, four points are extracted. These are the nearest surface points to the four robot joints under consideration. Each of the four cluster points is then transformed into the master robot’s base frame in which the RL state space is defined. 

\textbf{Fixed-base collaborative agents.}
\begin{figure*}
    \centering
    \includegraphics[width=0.8\textwidth]{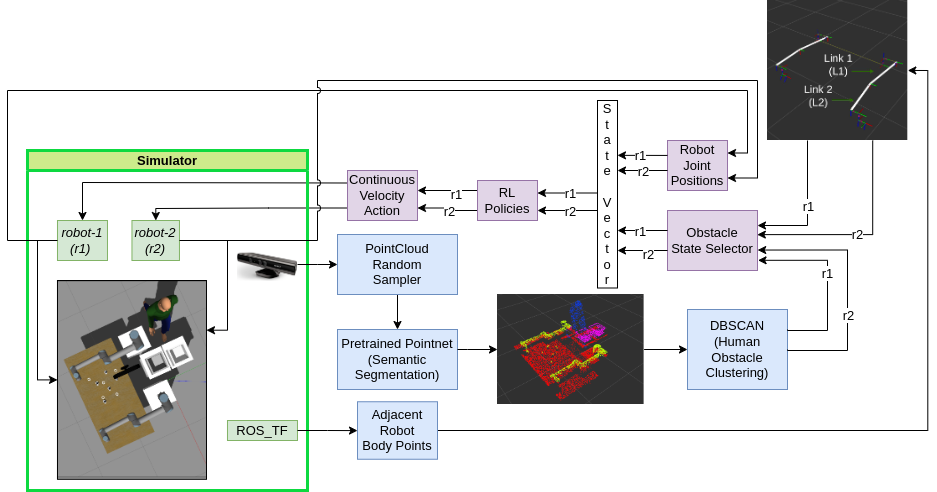}
    \caption{MLDEnv pipeline}
    \label{fig:pipeline}
\end{figure*}
We use \textit{ROS-TF} to extract the body points of the robots. The joint positions are published by \textit{ROS-TF} and the longest two links ($L_1$ and $L_2$ in Fig.~\ref{fig:pipeline}) of the manipulators are reconstructed with 100 uniformly sampled points assuming the links are the central line between two joints. These central body points are then used as obstacle points to each other master robot. From the sampled points on the obstacle robot body, four were extracted per body link, the ones that are closest to 
the selected four joints of the master robot. The distance of the master robot joints to obstacles is used in building the state space of the RL agent.

\subsection{Policy learning}
The problem of collaborative multi-agent pick-and-place from a shared and dynamic workspace can be modeled as Markov Decision Processes (MDP). The states of a single process are a combination of one agent or manipulator’s joint positions and the surface point coordinates extracted from the clustered obstacles (static and dynamic). The actions of a single process are joint velocity commands. The reward function is a piece-wise function based on distances from obstacles and the goal. Hence, we denote the observable state of an RL agent \enquote*{i} at time t as $\mathbf{s}_t^i = [ \theta_t^{i,j}, \mathbf{p}_t^{so}, \mathbf{p}_t^{do} ] $ and the action of the RL agent \enquote*{i} at time t as $\mathbf{a}_t^i = [ \dot \theta_t^{i,j}]$. $j= \{ 1..6 \}$ corresponds to the agent joints, $\mathbf{p}= [p_x, p_y, p_z]$ represents the position of \enquote*{so} static obstacles and \enquote*{do} dynamic obstacles. We model each robot with a separate MDP. Hence, in this work, we train two RL agents.
Each agent is trained with an on-policy gradient method, the PPO (Proximal Policy Optimization), a class of reinforcement learning algorithms~\cite{schulman2017proximal}. PPO achieves in general good learning performance, having the advantages of implementation simplicity (for its first-order objective function approximation), training stability, and sample efficiency. PPO trains an actor network to learn an optimal policy $\pi^*$, i.e. a mapping between states and actions, based on which a critic network evaluates the actions taken by updating a learnable state-action value function. Future works will consider a performance comparison of different stochastic methods in learning the collaborative pick-and-place task.

\subsubsection{Reward Function}
In state $\mathbf{s}_t$ at time t, the RL agent samples and
executes action $\mathbf{a}_t$ according to its policy $\pi(\mathbf{a}_t |\mathbf{s}_t )$, it then transitions to a new state $\mathbf{s}'_t$ according to the dynamics $p(\mathbf{s}'_t |\mathbf{s}_t , \mathbf{a}_t)$ and receives a reward $R(\mathbf{s}_t , \mathbf{s}'_t, \mathbf{a}_t )$. We consider finite-horizon discounted return problems, where the cumulative return at time t is the $\gamma$-discounted future return from time t to T (length of an episode), given by $G_t = \sum_{k=1}^{T} \gamma^{k}R_{t+k}$. The goal is to find the optimal policy $\pi^*$ which maximizes the expected sum of returns, given by the state-action value function $V_{\mathbf{a}_t\thicksim\pi}(\mathbf{s}_t, \mathbf{a}_t ) = \mathop{\mathbb{E}}_\pi [G_t|\mathbf{s}_0 = \mathbf{s}_t, \mathbf{a}_0 = \mathbf{a}_t]$. The reward function, $R$, is a piece-wise function representing the desired RL agent performance by ensuring the agent gets a higher reward when performing desired actions and a lower reward in the other cases. There are two parameters in the reward function defined in Eq.~\eqref{eq:reward}. Those are the distance from the end-effector to the goal ($\emph{d}_{eg}$) and the distance of the nearest obstacle from each master robot joint under consideration (\emph{$d_1$, $d_2$, $d_3$, $d_4$}). There are also pre-assigned weights for each of the distances (\emph{$w_1$, $w_2$, $w_3$, $w_4$}). These weights are assigned to joints based on their location in the manipulator. A virtual sphere (\emph{S}) of a radius of 40cm is created for safety measures around the four robot joints. Any obstacle inside this sphere is considered to be in danger. Both robots have their own goal positions in our case study and since robots are considered as obstacles to each other, the radius of the sphere was selected based on the distance between the goals position of the robots. Whenever an obstacle (\emph{O}) is inside any of the four spheres, a weighted sum of the distances from the nearest obstacle point to each robot joint is considered for penalizing the policy. 

\begin{equation} \label{eq:reward} R = \begin{cases} 
      2* D_{min} - l_1 -(d_{eg}/2) & \exists O  \in S\\
      -(d_{eg}/2) & \forall O \notin S \\
      2 & d_{eg} \leq l_2
   \end{cases}
\end{equation}
\[D_{min} = w_1\cdot d_1 + w_2 \cdot d_2 + w_3 \cdot d_3 + w_4 \cdot d_4\]
and where the hyper-parameters $l_1$ and $l_2$ are set to 5 and 15cm, respectively. The reward function is designed to provide a higher reward when the agent moves simultaneously away from the obstacles and towards the goal, while a lower reward is assigned when the agent moves near the obstacles.

\subsubsection{Policy}
The policy is learned using a multi-layer perceptron. The problem statement deals with a variable number of obstacles. To accommodate this dynamic variation, an LSTM-based obstacle embedding technique has been adopted from \cite{everett2018motion}. Each obstacle has four nearest points, each to one of the master agent joints, from which a weighted sum distance $D_{min}$ is calculated. The LSTM layer takes in each obstacle’s surface points in the order of their corresponding weighted sum distance to the master agent. The output of the LSTM layer is concatenated with the master robot joints' position in radians. Leveraging a PPO model, in addition to the actor network, another similar neural network was used as a critic to estimate the value function and subsequently evaluate the expected cumulative reward at the next state. Both policy and value functions are trained simultaneously with PPO.

\section{Experiments} \label{sec:exper}
\subsection{PointNet Evaluation}
\begin{figure}
    \centering
    \begin{subfigure}[b]{0.2\textwidth}
    \centering
    \includegraphics[width= 0.5\textwidth, trim={5cm 2cm 5cm 2cm} , clip]{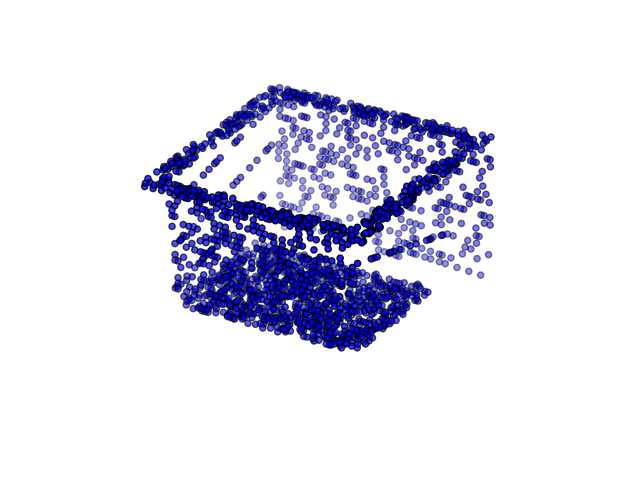}
    \caption{Box}
    \end{subfigure}
        \begin{subfigure}[b]{0.2\textwidth}
    \centering
\includegraphics[width=0.5\textwidth, trim={5cm 7cm 5cm 5cm} , clip]{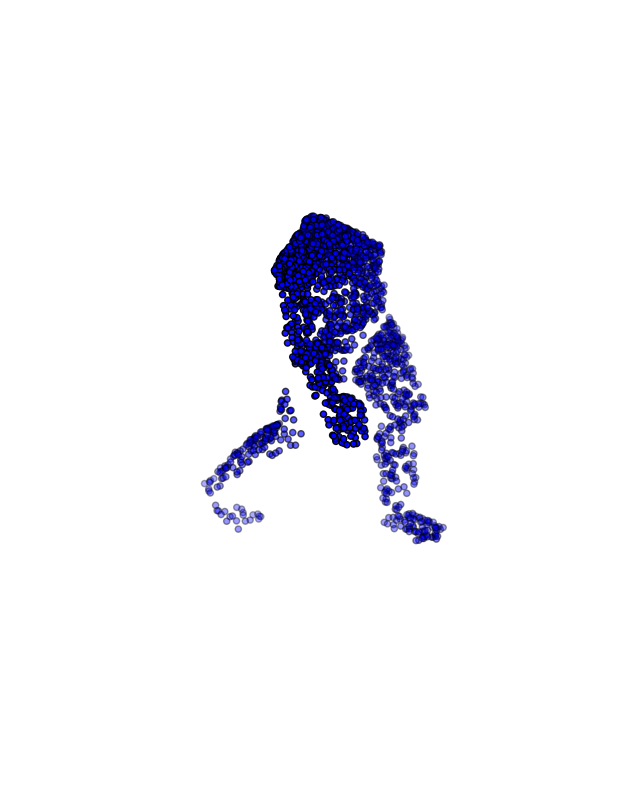}
    \caption{Human}
    \end{subfigure}
        \begin{subfigure}[b]{0.2\textwidth}
    \centering
    \includegraphics[width=0.5\textwidth, trim={5cm 2cm 5cm 2cm} , clip]{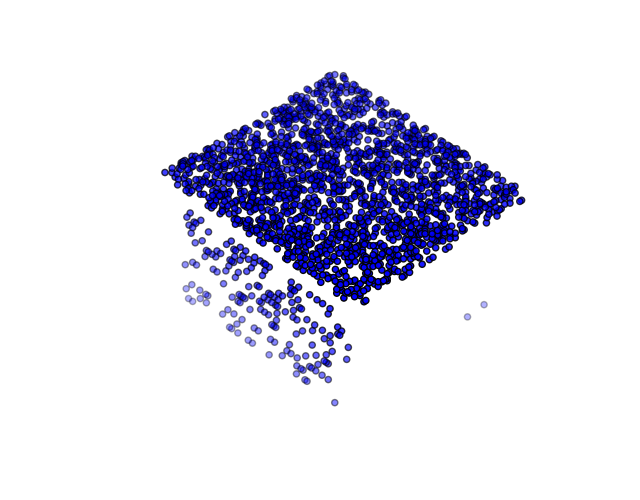}
    \caption{Table}
    \end{subfigure}
        \begin{subfigure}[b]{0.2\textwidth}
    \centering
    \includegraphics[width=0.5\textwidth, trim={5cm 2cm 5cm 2cm} , clip]{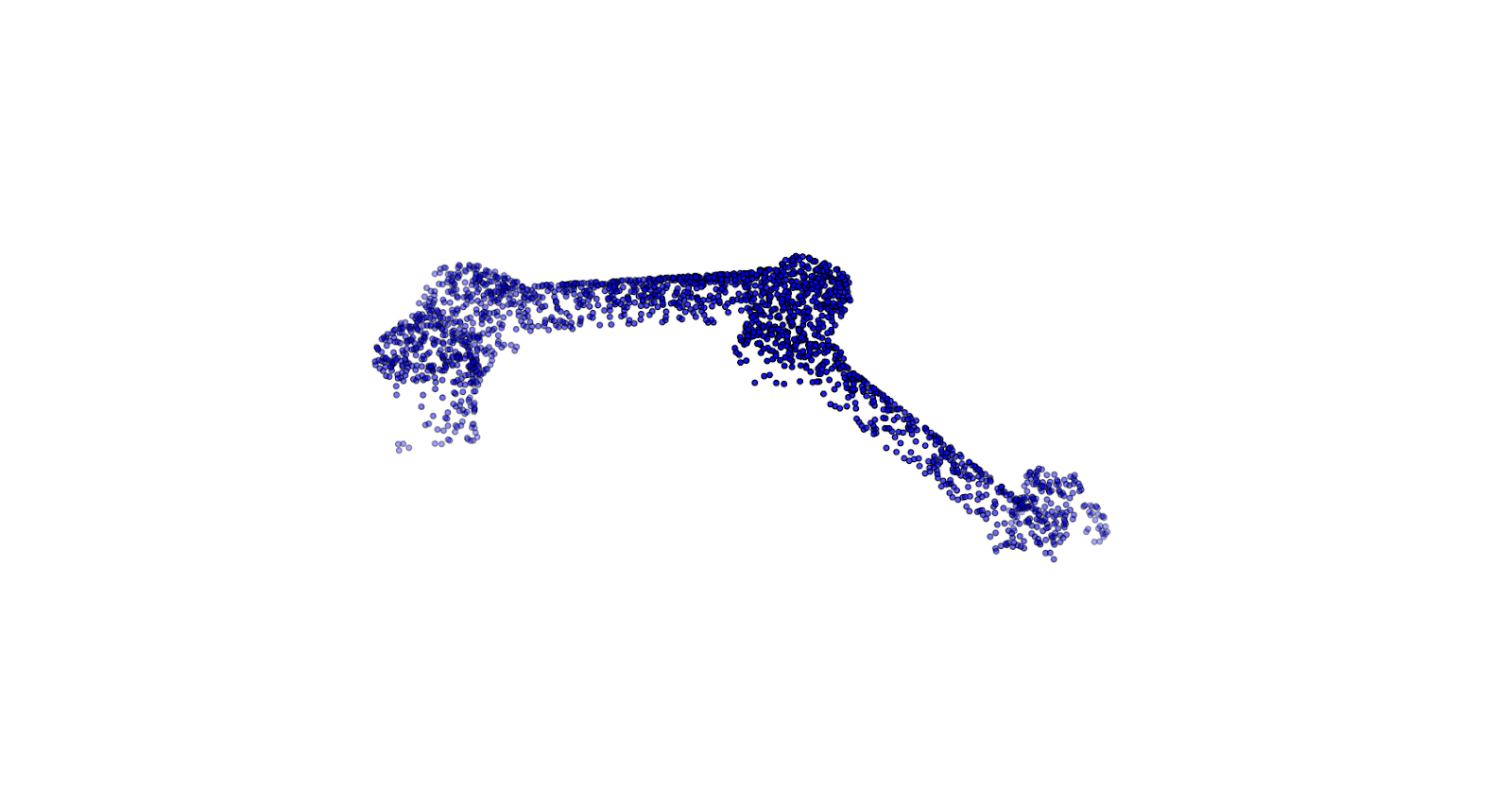}
    \caption{Manipulator}
    \end{subfigure}

    \caption{Sample data for PointNet training: box, human, table, robot.}
    \label{fig:samples}
\end{figure}

\begin{figure}[htb]
\begin{minipage}{0.5\textwidth}
\begin{tikzpicture}
  \node (img) { 
   \includegraphics[width=7cm, trim ={0 0 0 0.7cm} , clip]{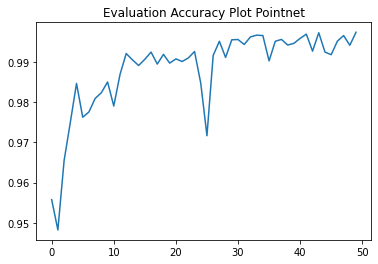}
} ;
  \node[below=of img, node distance=0cm, yshift=1.2cm,font=\color{black}] {epochs};
  \node[left=of img, node distance=0cm, rotate=90, anchor=center,yshift=-0.95cm,font=\color{black}] {accuracy};
    \label{fig:pointnet_testing}
 \end{tikzpicture}
\end{minipage}%
%
    

\caption{Evaluation accuracy of PointNet on custom dataset}
    \label{fig:pointnet}
\end{figure}

The PointNet was trained on a dataset collected in the simulation with a simulated Kinect RGB-D sensor. 
Each object in a given scene configuration was individually captured at a different orientation and the resulting point clouds of every object in a single scene, as in Fig.~\ref{fig:samples}, were combined to produce a new annotated scene. The PointNet was trained for 50 epochs and with a batch size of 8. The evaluation accuracy plot can be seen in Fig.~\ref{fig:pointnet} where a segmentation accuracy close to 1.0 is achieved.

\subsection{Training a Reinforcement Learning Agent}
The reinforcement learning agent was trained with the Proximal Policy Optimization (PPO) algorithm. Each manipulator in the scenario was controlled with an RL agent. An OpenAI Gym environment was created to set the learning process of the manipulator with PPO. The human in the scene was continuously moving to simulate a dynamic obstacle for the manipulators. 

\begin{figure}[H]
    \centering
    \includegraphics[width=7cm]{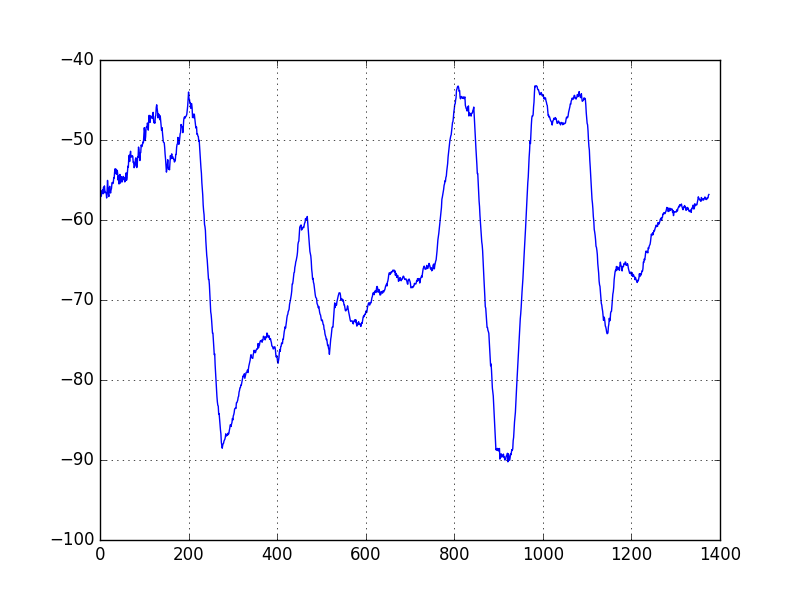}
    \caption{Robot-1 cumulative reward over 40 steps episode (a total of 1400 episodes), with human-in-the-scene.}
    \label{fig:robot1 reward with human}
\end{figure}

\begin{figure}[htb]
    \centering
    \includegraphics[width=7cm]{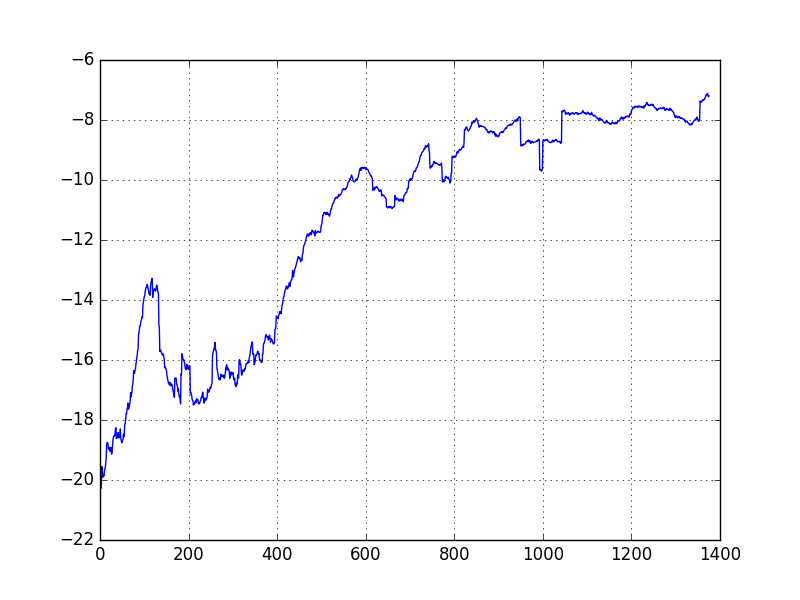}
    \caption{Robot-2 cumulative reward over 40 steps episode (a total of 1400 episodes), with human-in-the-scene.}
    \label{fig:robot2 reward with human in the scene}
\end{figure}

\textbf{Experimental evaluation with human-in-the-scene.}
This experiment setting was done with humans in the scene, considering a more realistic human-robot co-existence scenario. Both agents were trained for 56,000 timesteps. The plots in Figs.~\ref{fig:robot1 reward with human} and~\ref{fig:robot2 reward with human in the scene} show that the agent farther from the human (\textit{robot-2} in Fig.~\ref{fig:robot2 reward with human in the scene}) was able to learn the objective quicker than that of the agent closer to the human (Fig.~\ref{fig:robot1 reward with human}).

\FloatBarrier

\section{Conclusion and Future Directions} \label{sec:conc}
The work presents a two-stage framework for safe collaborative robots deployed in industrial pick-and-place operations. It leverages an actor-critic deep reinforcement learning framework to learn a stochastic policy in a dynamic and shared workspace between human operators and dual robots. Future work will be devoted to extending MLDEnv framework to real-world policy deployment by first exploring the advantage of imitation learning, as the probabilistic movement primitives, for a fast policy convergence, and second, exploring probabilistic pose estimator for the obstacles in the environment.
\bibliographystyle{IEEEtran}
\bibliography{refs.bib}

\end{document}